\documentclass{article}

\usepackage[preprint]{corl_2026}

\usepackage{booktabs}
\usepackage{amsfonts}
\usepackage{amsmath}
\usepackage{graphicx}
\usepackage{pifont}
\usepackage{tabularx}
\usepackage{multirow}
\usepackage{wrapfig}

\definecolor{resireggray}{gray}{0.5}

\newcommand{\graycell}[1]{{\color{resireggray}#1}}

\title{ReSiReg: Towards Spatially Consistent Semantics\\in Language-Conditioned Robotic Tasks}

\author{
  Simon Schwaiger\textsuperscript{1,2}
  \quad
  David Seyser\textsuperscript{2}
  \quad 
  Alessandro Scherl\textsuperscript{2,3}  
  \\
  \textbf{Wilfried Wöber\textsuperscript{4}}
  \quad
  \textbf{Gerald Steinbauer-Wagner\textsuperscript{1}}
  \\[3pt]
  \textsuperscript{1}Graz University of Technology, Institute of Software Engineering and Artificial Intelligence
  \\[3pt]
  \textsuperscript{2}University of Applied Sciences Technikum Wien, Department of Industrial Engineering
  \\[3pt]
  \textsuperscript{3}University of Alicante, Department of Computer Technology
  \\[3pt]
  \textsuperscript{4}University of Natural Resources and Life Sciences,\\Institute for Integrative Nature Conservation Research
  \vspace{-9mm}
}

\begin{document}


\makeatletter
\newcommand{\thickhline}{%
    \noalign {\ifnum 0=`}\fi \hrule height 1pt
    \futurelet \reserved@a \@xhline
}
\newcolumntype{"}{@{\hskip\tabcolsep\vrule width 1pt\hskip\tabcolsep}}
\newcolumntype{'}{@{\hskip\tabcolsep\vrule width 0.5pt\hskip\tabcolsep}}
\newcolumntype{[}{@{\vrule width 1pt\hspace{6pt}}} \newcolumntype{]}{@{\hspace{6pt}\vrule width 1pt}}


\maketitle


\begin{abstract}
Vision-Language Models (VLMs) enable robots to follow open-language instructions.
However, dense VLM embeddings have shown to be noisy and lack spatial consistency.
This is problematic for robotic applications, which require simultaneous reasoning over semantics and 3D space.
We examine spatial structure across recent VLMs and propose ReSiReg, a feature reconstruction method that uses spatially consistent VLM intermediates to improve dense language-grounded retrieval.
ReSiReg clusters intermediates into visual prototypes, derives their language descriptors, and reconstructs each patch as a soft mixture of prototype-level language embeddings.
We evaluate quantitatively on OVSS and 3D mapping across backbones, and qualitatively in real-world manipulation scenes. Quantitative results show improved dense retrieval; manipulation scenes show more spatially consistent target activations.
We further provide a compact 25M dense VLM for robotic applications, substantially smaller than and competitive with ViT-B baselines.
Available at \url{https://resireg.github.io}

\end{abstract}



\begin{figure}[!h]
    \centering
    \includegraphics[width=0.99\linewidth]{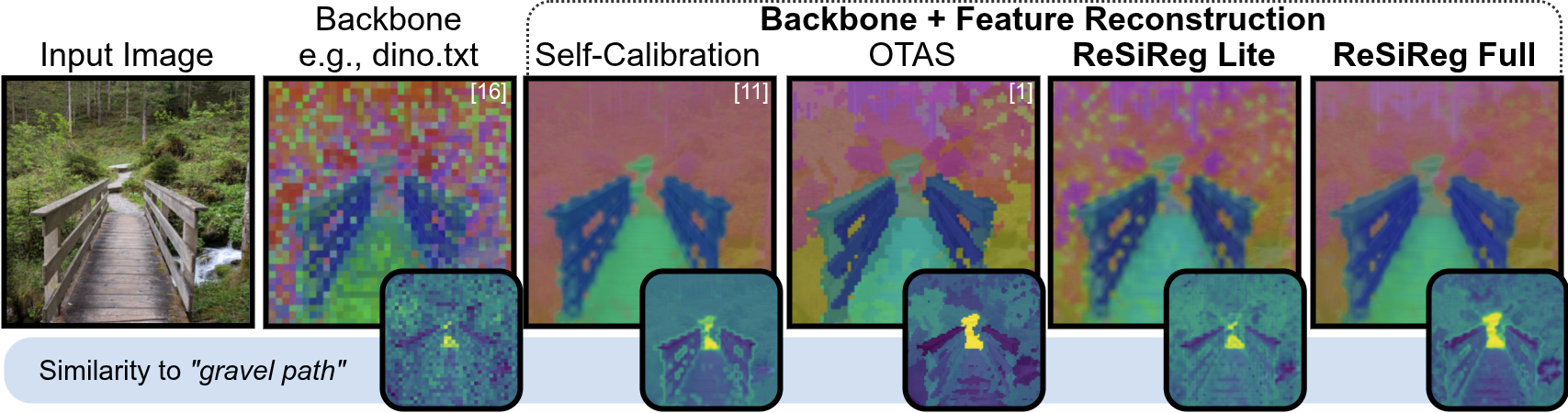}
    \vspace{-3mm}
    \caption{\textbf{ReSiReg} is a feature reconstruction method for language-grounded backbones. It recovers spatially consistent dense embeddings, even under heavy view-dependent noise. Top: PCA over backbone with feature reconstruction methods. Bottom: Similarity to \textit{"gravel path"}.}
    \vspace{-2mm}
    \label{fig:teaser}
\end{figure}

\section{Introduction}
\label{sec:intro}

Language representations have shown to benefit robot applications due to the perception of abstract concepts and fuzzy semantic boundaries \cite{Schwaiger2025OTAS}, and ability to condition exploration, mapping \cite{Alama2025Rayfronts}, and manipulation on semantic targets \cite{huang2023voxposer}.
Language-grounded robot control typically relies on vision-language encoders supervised by contrastive objectives over image-caption pairs \cite{Radford2021CLIP, Cherti2023OpenCLIP}.
During training, language-grounding is incentivised on the CLS token, while dense patch-level features are extracted using backbone-specific modifications such as distribution shifts \cite{Zhou2021MaskCLIP2}, post-training, \cite{Wysoczanska2025CLIPDinoiser}, or alternative feature projections \cite{Alama2025Rayfronts, alama2025radseg}.
However, for robotic applications, a problem emerges: Vision Language Models (VLMs) lack spatial consistency \cite{Schwaiger2025OTAS, Qiu2024Featuresplatting}, resulting in noisy dense predictions (see Fig.~\ref{fig:teaser}).
This is detrimental to robotic systems, which must jointly reason over semantics and 3D space to complete tasks.
Methods have attempted to improve spatial consistency through self-similarity \cite{Wang2025SClip, alama2025radseg}, anomaly filtering \cite{Bai2025SelfCalibratedCLIP}, and conditioning on other (typically self-supervised, SSL) foundation models \cite{Schwaiger2025OTAS, Wysoczanska2025CLIPDinoiser, Shi2025Trident, Lan2025ProxyCLIP}.
However, these methods are backbone-specific without examining applicability to other VLMs, and add computational complexity through other backbones.
Recent works on VLMs have shown model intermediates that are spatially consistent, but not language-grounded.
These intermediary representations are either explicitly incentivised during training \cite{Ranzinger2024AMRADIO, Heinrich2024RADIO2, Jose2025DinoTXT}, or form naturally before the final language-grounded projections \cite{Bai2025SelfCalibratedCLIP, Bolya2025PerceptionEncoder}.
We hypothesise that this property can be leveraged to apply the spatially consistent structure of model intermediates to the final language-grounded embeddings.

\begin{figure}[!t]
    \centering
    \includegraphics[width=0.94\linewidth]{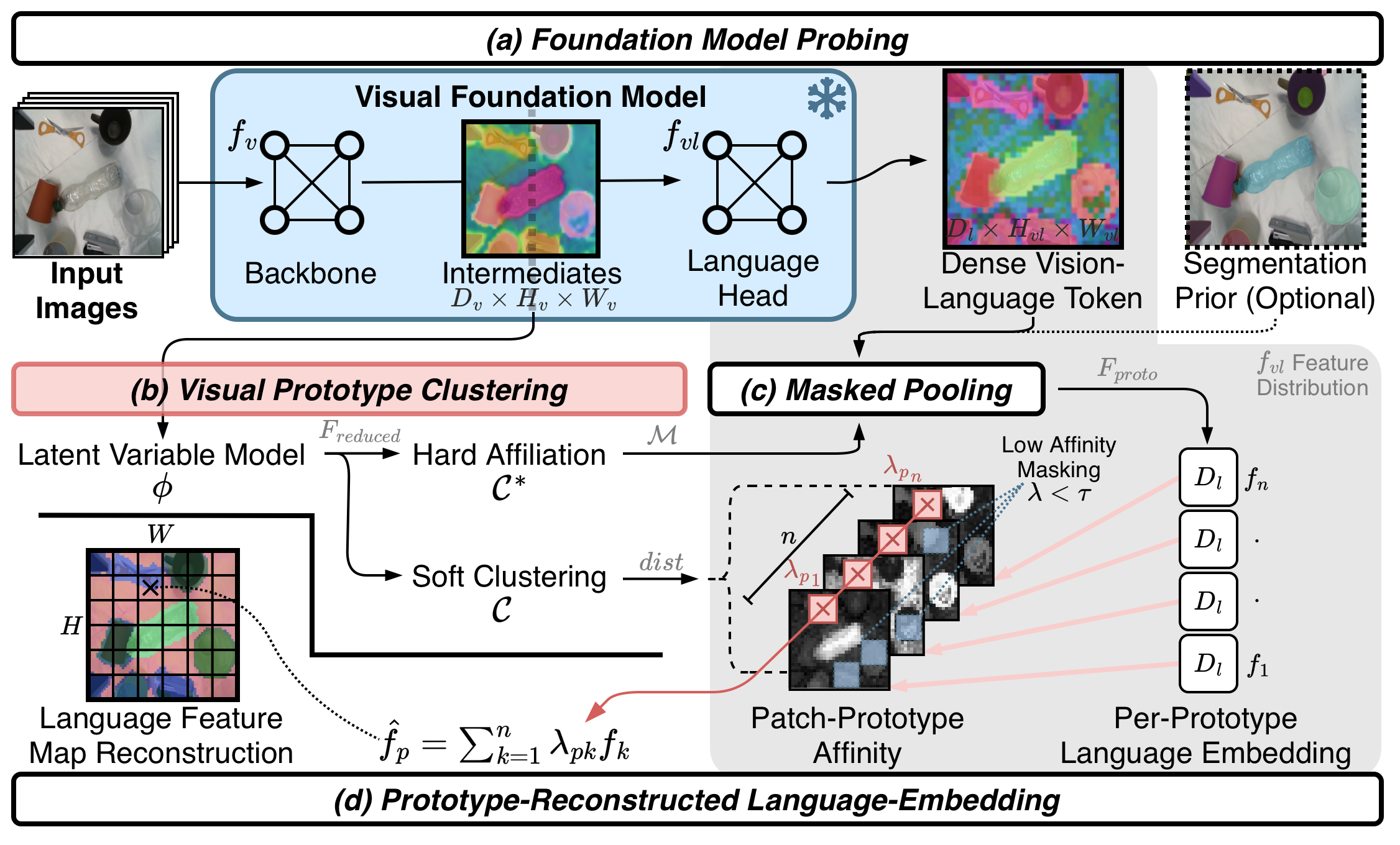}
    \vspace{-4mm}
    \caption{\textbf{ReSiReg Feature Reconstruction.} (a) Foundation model intermediates, language-aligned output tokens, and an optional segmentation prior are aggregated. (b) Intermediate tokens are then decorrelated and reduced in dimensionality through a latent-variable model and clustered to visual prototypes. (c)  Masked pooling is applied over hard clusters, language tokens, and the optional segmentation prior to determine language embeddings representing each visual prototype. (d) Finally, per-patch features are aggregated as linear combinations of all visual prototypes.}
    \label{fig:method_overview}
    \vspace{-2mm}
\end{figure}

We present a method for \textbf{Resi}dual feature \textbf{Re}construction and language \textbf{G}rounding (ReSiReg).
ReSiReg retrieves spatially consistent VLM embeddings, a property previously attributed to SSL \cite{Oquab2023DINOv2, Darcet2023VitRegisters, Simeoni2025Dinov3} and distilled backbones \cite{Ranzinger2024AMRADIO, Heinrich2024RADIO2}.
Similarly to \cite{Schwaiger2025OTAS}, ReSiReg first decomposes VLM intermediates into visual prototypes, disentangling the spatial structure of semantic concepts observed in the scene.
Motivated by explanatory factor analysis, which shows that models implicitly learn mixtures of semantic concepts \cite{woeber2023nonlinear}, we reconstruct a language-grounded and spatially consistent scene representation from mixtures of visual prototypes on image- up to scene-level.
The resulting method acts as a post-hoc residual branch on existing pretrained VLMs that conditions language-grounded outputs on the spatially consistent structure of model intermediates. 
To summarise, our contributions are:
\begin{itemize}
    \item a feature reconstruction method that improves dense semantic retrieval from spatially consistent VLM intermediates without increasing model parameters, and
    \item empirical grounding across recent VLM backbones and robotics-relevant retrieval. 
    Based on our findings, we provide a 25M parameter VLM, enabling high throughput and spatially consistent dense output; both capabilities integral to language-driven robotic applications.
\end{itemize}
We demonstrate our method using 2D Open-Vocabulary Semantic Segmentation (OVSS), 3D semantic mapping, and open-language manipulation across multiple recent VLM backbones.
The results show improved dense semantic retrieval, while retaining real-time online performance.


\section{Related Work}
\label{sec:sota}

VLMs project images and text into a joint feature space, where semantic concepts are embedded adjacently across modalities \cite{Radford2021CLIP, Cherti2023OpenCLIP}.
To ground model outputs in language, backbone-specific modifications have been introduced \cite{Zhou2021MaskCLIP2, Alama2025Rayfronts, alama2025radseg}.
However, view-dependent noise in language embeddings degrades dense predictions \cite{Schwaiger2025OTAS, Bai2025SelfCalibratedCLIP, Qiu2024Featuresplatting} (see Fig.~\ref{fig:teaser}).
In robotics, an intuitive way to improve spatial consistency is fusing multiple observations over time to semantic maps \cite{Alama2025Rayfronts, alama2025radseg, Qiu2024Featuresplatting, Kerr2023LERF, Yamazaki2024OpenFusion}.
This, however, requires multiple views from different angles to be effective, requires intrinsic and extrinsic camera calibration, and introduces computational overhead.

Backbone modifications have been proposed to reduce view-dependent noise at encoder-level.
\cite{alama2025radseg, Shi2025Trident, Lan2025ProxyCLIP} aggregate semantically similar structures in the language-grounded feature map.
\cite{Schwaiger2025OTAS, Qiu2024Featuresplatting} ground dense language features through hard masks, improving spatial consistency at the cost of discretising patch-level embeddings.
\cite{Alama2025Rayfronts, alama2025radseg, Hajimiri2025NaClip} modify backbone attention mechanisms.
\cite{Bai2025SelfCalibratedCLIP} propose anomaly detection to identify and subsequently prune dense outputs contributing to noise in CLIP feature maps.
While effective to varying degrees, these modifications have to be carefully tuned to each backbone with limited applicability across backbones or even model sizes of the same backbone.

Alternative methods propose conditioning VLM outputs on other backbones to improve dense semantic structure \cite{Schwaiger2025OTAS, Wysoczanska2025CLIPDinoiser, Shi2025Trident, Lan2025ProxyCLIP}.
Mainly SSL models, such as the DINO family of models \cite{Oquab2023DINOv2, Darcet2023VitRegisters, Simeoni2025Dinov3} are used due their high feature consistency across views and scaling to scene-level representations \cite{man2024lexicon3d}.
This, however, adds computational complexity for auxiliary backbone inference and the grounding mechanism.
Motivated by the semantic structure of SSL models, \cite{Cao2026TIPSv2} propose to modify contrastive training for language-grounding with an additional iBot-style SLL loss \cite{Zhou2022iBOT}.

While existing methods reduce dense VLM prediction noise, they are 1) backbone-specific with limited applicability across backbones \cite{alama2025radseg, Zhou2021MaskCLIP2, Alama2025Rayfronts, Shi2025Trident, Lan2025ProxyCLIP, Hajimiri2025NaClip}, 2) rely on additional backbones \cite{Schwaiger2025OTAS, Wysoczanska2025CLIPDinoiser, Shi2025Trident, Lan2025ProxyCLIP}, or 3) rely on heavy large-scale training procedures \cite{Cao2026TIPSv2}.
In contrast, this paper proposes a middle ground, built on the observation that VLM intermediates already observe a spatially consistent dense structure \cite{Wang2025SClip, Ranzinger2024AMRADIO, Heinrich2024RADIO2, Jose2025DinoTXT, Bolya2025PerceptionEncoder}.
We introduce a feature reconstruction mechanism that conditions language-grounded outputs on these VLM intermediates.
The method differs from prior work both by reconstructing dense outputs through soft prototype affinities and by applying this reconstruction as a residual post-hoc improvement across VLM backbones.


\section{Method}
\label{sec:method}

Fig.~\ref{fig:method_overview} presents an overview of ReSiReg.
Language-grounded backbones are probed to extract spatially-consistent intermediary tokens and dense language-grounded final embedding (a).
Intermediates are clustered, resulting in hard clusters and soft cluster distances (b). Hard clusters are concatenated with an optional segmentation prior and pooled over to obtain a language embedding for each visual prototype in the scene (c).
Soft cluster affinities determine a linear combination of prototypes to reconstruct each patch of the spatially consistent, language-grounded feature map (d).

\subsection{Foundation Model Probing}
Given an input image $I \in \mathbb{R}^{H \times W \times 3}$, the goal is to extract spatially consistent backbone intermediates $F_v = \mathbf{f}_v(I) \in \mathbb{R}^{D_v \times H_v \times W_v}$ and dense language-aligned tokens $F_{vl} = \mathbf{f}_{vl}(F_v)  \in \mathbb{R}^{D_l \times H_{vl} \times W_{vl}}$ from the frozen foundation models.
Backbone is denoted as $\mathbf{f_v}$ with language head $\mathbf{f_{vl}}$.
$H$ and $W$ denote height and width of the original image ($H, W$), backbone patch token ($H_v, W_v$), and language patch token ($H_{vl}, W_{vl}$) respectively.
$\times$ denotes spatial tensor dimensions.

Due to contrastive training, typically only the CLS token of Vision-Transformer (ViT)-based VLMs is language grounded \cite{Zhou2021MaskCLIP2}.
For each frozen backbone, we extract dense language-grounded features using the dense projection mechanism of the model family: MaskCLIP-style value-path projection for CLIP \cite{Zhou2021MaskCLIP2}, swapping patch- and CLS adapter heads for RADIO \cite{Alama2025Rayfronts, alama2025radseg}, and the trained predictor for dino.txt \cite{Jose2025DinoTXT}.
Intermediates are probed from the models, since they observe increased spatial consistency but lack language-grounding.
CLIP models follow \cite{Wang2025SClip}.
Models with post-trained language heads are probed before the head, and agglomerative models are probed from the shared backbone.

\subsection{Visual Prototype Clustering}
To condition the language-grounded embeddings on the intermediate semantic structure, $F_v$ embeddings are decomposed to visual prototypes through clustering.
Visual prototypes denote the latent semantic scene structure learned by a foundation model.
Clustering embeddings into visual prototypes has been shown to transfer spatial feature structure onto language-grounded feature maps \cite{Schwaiger2025OTAS}.
Prior methods aggregate language features over hard masks, which discretises the resulting feature map and assumes that clusters align with downstream target classes.
This can suppress fine-grained object structure and continuous semantic transitions.
ReSiReg retains hard assignments for language-grounding, while using soft affinities for reconstruction.
It is further differentiated by building clusters on intermediates rather than computationally expensive extra backbones.
For readability, mathematical notation depicts a single image and omits the batch index.

We therefore extend previous hard-mask grounding methods \cite{Schwaiger2025OTAS, Qiu2024Featuresplatting} with same-backbone residual reconstruction over soft prototype affinities.
Intermediates $F_v$ are first interpolated to a shared spatial grid using bilinear interpolation $U_{bl}$ and a scaling factor $s$, then flattened: $F_v^{flat} = U_{bl}(F_v) \in \mathbb{R}^{D_v \times H_v \cdot W_v \cdot s^2}$.
Flattened features are decorrelated and reduced in embedding dimension using a latent-variable model (LVM) with subsequent normalisation.
$F_{reduced} = \mathrm{norm}_2\!\left(\phi(F_v^{flat})\right) \in \mathbb{R}^{r \times H_v \cdot W_v \cdot s^2}$, where $r$ is the reduced dimension and $\mathrm{norm}_2$ denotes $\ell_2$ normalisation along the feature axis.
Soft Clustering $\mathcal{C}$ is applied to decompose $F_{reduced}$ to $n$ prototype centroids $\mu \in \mathbb{R}^{r \times n}$.
For patch $p$ and prototype $k$, the distance map is defined as $d_{pk}=\left\lVert z_p-\mu_k\right\rVert_2$, where $z_p$ is the corresponding column of $F_{reduced}$ and $d\in\mathbb{R}^{H_v \cdot W_v \cdot s^2 \times n}$.
Downstream, the distances are converted to semantic affinities between all patches and prototypes for dense feature reconstruction.

To enable subsequent language grounding of visual prototypes, feature aggregation over masks is required.
Therefore, soft clusters are discretised to hard clusters using $C^* = \operatorname*{argmin}_{k}(d_{pk})$ and reshaped to a mask depicting the closest affiliation between each patch and centroid $\mathcal{M} \in \mathbb{R}^{H_v \cdot s \times W_v \cdot s}$.

\subsection{Masked Pooling}

To ground intermediate structure in language, masked pooling is applied over closest prototype affiliation $\mathcal{M}$ and language embedding $F_{vl}$.
Contrary to prior work, which retrieves a discretised feature map from this step, we apply masked pooling to obtain a language embedding for each visual prototype.
Clustering yields a distance map $d_{pk}$ between each patch $p$ and visual prototype $k$, and the closest prototype affiliation $\mathcal{M}$.
The mask defines the prototype support as $\Omega_k=\{p\mid \mathcal{M}_p=k\}$, with $\Omega_k$ as the set of patches assigned to prototype $k$.
Dense language features are aligned to the mask grid and flattened as $L=\mathrm{vec}\!\left(\mathcal{U}_{nn}(F_{vl})\right)$, where $L_p\in\mathbb{R}^{D_l}$ denotes the language token at patch $p$.
Prototype-level embeddings $f_k$ are obtained by masked average pooling over $\Omega_k$.

The mask and language-feature grids may have different spatial resolutions.
Therefore, dense language features are aligned to the mask grid using nearest-neighbour interpolation only at the patch level, avoiding near pixel-level language tensors.
This mechanism can incorporate an optional segmentation prior $S$ with instance labels $S_p\in\{0,\dots,m\}$.
Each prior instance $j\in\{1,\dots,m\}$ defines an additional prototype support $\Omega_{n+j}=\{p\mid S_p=j\}$ and hard prior distance channel $d^{S}_{p,n+j}=1-\mathbf{1}[S_p=j]$, which is concatenated with the cluster-induced distances before affinity computation and reconstruction.
Let $\tilde{n}=n+m$ when the prior is used, and $\tilde{n}=n$ otherwise.
From $\Omega$, language embeddings are constructed as normalised averages over each non-empty prototype.
\begin{equation}
    f_k=\mathrm{norm}_2\!\left(\frac{1}{|\Omega_k|}\sum_{p\in\Omega_k}L_p\right),
    \qquad \forall k\in\{1,\dots,\tilde{n}\}:|\Omega_k|>0.
\end{equation}
For empty supports, $f_k$ is a zero descriptor without semantic contribution.
Stacking prototypes yields $F_{\mathrm{proto}}=[f_1,\dots,f_{\tilde{n}}]^{\top}\in\mathbb{R}^{\tilde{n}\times D_l}$ where each $f_k$ is a language-grounded descriptor for visual prototype $k$.
$F_{\mathrm{proto}}$ are subsequently reconstructed to a spatially consistent language-representation.

\subsection{Prototype-Reconstructed Language Embedding}

\cite{woeber2023nonlinear} show that visual encoders implicitly learn mixtures of latent concepts during training.
This can be leveraged to condition one feature distribution on the structure of another model.
Intuitively, broadcasting each $f_k$ from masked pooling to patches $p\in\Omega_k$ yields a feature map in $f_{vl}$ feature distribution, conditioned on visual prototypes of $f_v$, following \cite{Schwaiger2025OTAS}.
However, this discretises the reconstructed feature map and assumes clusters align with downstream target classes.
Instead, our goal is to preserve smooth and fine-grained structure in output embeddings.
Therefore, to reconstruct a dense map in $\mathbb{R}^{D_l}$, distances $d_{pk}$ are converted to nonnegative affinities using a temperature-scaled softmax, with the resulting $\tilde{\lambda}_{pk}$ representing the affinity of patch $p$ to visual prototype $k$.
\begin{equation}
    \tilde{\lambda}_{pk}
    =\frac{\exp(-\alpha d_{pk})}
    {\sum_{j=1}^{n}\exp(-\alpha d_{pj})},
    \qquad \alpha>0
\end{equation}

\textbf{Low Affinity Masking.}
We adopt affinity gating \cite{alama2025radseg, Lan2025ProxyCLIP} below a cutoff $\tau\ge0$ using $\lambda_{pk}=\tilde{\lambda}_{pk}\,\mathbf{1}\!\left[\tilde{\lambda}_{pk}\ge\tau\right]$, where $\tau$ trades mixture diversity against attachment to high-confidence prototypes.
Since the surviving weights are not renormalised, $\sum_{k=1}^{n}\lambda_{pk}\le1$ after gating.

Since each dense output language feature is a mixture of prototypes, we retrieve the final per-patch language feature as a gated linear combination of prototype embeddings using $\hat{f}_p=\sum_{k=1}^{\tilde{n}}\lambda_{pk}f_k$.
The reconstructed features $\{\hat{f}_p\}$ are normalised and reshaped to the spatial layout of $\mathcal{M}$, yielding $\hat{F}_{vl}$.
The result is a language-grounded reconstructed feature map that transfers the spatial consistency of backbone intermediates into the $D_l$-dimensional output of $f_{vl}$.
This feature reconstruction also directly extends to batches of input images and 3D projection given camera extrinsics and intrinsics.
We show this method's benefit to multiple backbones and robotic tasks in our experiments.


\section{Experiments}
\label{sec:results}

\begin{table}[!ht]
    \centering
    \begin{tabular}{m{14mm} c" c c c c|c c c c}
        \thickhline
        & 
        & \multicolumn{4}{c}{\textbf{ADE20K} [mIoU$\uparrow$]} 
        & \multicolumn{4}{c}{\textbf{ORAD-3D} [mIoU$\uparrow$]} \\

        & & & & \multicolumn{2}{c}{\textbf{ReSiReg}} & & & \multicolumn{2}{c}{\textbf{ReSiReg}} \\

        {Backbone}
        & Param
        & BB
        & Calib
        & \textbf{Lite}
        & \textbf{Full}
        & BB
        & Calib
        & \textbf{Lite}
        & \textbf{Full}  \\

        \hline \hline
        & & \multicolumn{8}{c}{\textbf{ViT-S}} \\

        EUPE &  &  & \graycell{4.77} & \graycell{12.31} & \graycell{11.74} & & \graycell{4.87} & \graycell{25.93} & \graycell{24.45} \\
        \scriptsize{{\textbf{Our~Head}}} & \multirow{-2}{*}{25M} & \multirow{-2}{*}{\graycell{12.66}} & -7.89 & -0.35 & -0.92 & \multirow{-2}{*}{\graycell{19.85}} & -14.98 & +6.08 & +4.6 \\
        
        \hline \hline
        & & \multicolumn{8}{c}{\textbf{ViT-B}} \\
        OTAS\textsuperscript{*} & 103M & \graycell{15.35} & & & & \graycell{12.53} & & & \\
        TIPSv2 & 86M & \graycell{15.67} & & & & \graycell{15.30} & & & \\

        \hline
        
            &  &  & \graycell{13.47} & \graycell{11.75} & \graycell{12.89} & & \graycell{10.56} & \graycell{9.03} & \graycell{10.78} \\
        \multirow{-2}{*}{CLIP\textsuperscript{*}} & \multirow{-2}{*}{86M} & \multirow{-2}{*}{\graycell{10.61}} & +2.86 & +1.14 & +2.28 & \multirow{-2}{*}{\graycell{8.17}} & +2.39 & +0.86 & +2.61 \\
 
        \rowcolor{gray!10}
        RADIOv3 &  &  & \graycell{17.11} & \graycell{25.01} & \graycell{24.36} &  & \graycell{23.30} & \graycell{22.55} & \graycell{22.05} \\
        \rowcolor{gray!10}
        \scriptsize{CLIP Head} & \multirow{-2}{*}{113M} & \multirow{-2}{*}{\graycell{23.30}} & -6.19 & +1.71 & +1.06 & \multirow{-2}{*}{\graycell{17.19}} & +6.11 & +5.36 & +4.68 \\
 
        RADIOv3 &  &  & \graycell{19.26} & \graycell{28.42} & \graycell{27.78} &  & \graycell{22.93} & \graycell{18.31} & \graycell{21.32} \\
        \scriptsize{{SigLIP~2~Head}} & \multirow{-2}{*}{113M} & \multirow{-2}{*}{\graycell{26.52}} & -7.26 & +1.90 & +1.26 & \multirow{-2}{*}{\graycell{16.80}} & +6.13 & +1.51 & +4.52 \\

        \rowcolor{gray!10}
        & & & \graycell{24.63} & \graycell{28.84} & \graycell{29.00} & & \graycell{16.02} & \graycell{24.21} & \graycell{25.00} \\
        \rowcolor{gray!10}
        \multirow{-2}{*}{RadSeg} & \multirow{-2}{*}{113M} & \multirow{-2}{*}{\graycell{28.83}} & -4.2 & +0.01 & +0.17 & \multirow{-2}{*}{\graycell{22.17}} & -6.15 & +2.04 & +2.83 \\

        \hline \hline
        & & \multicolumn{8}{c}{\textbf{ViT-L}} \\

        TIPSv2 & 304M & \graycell{19.26} & & & & \graycell{13.67} & & & \\

        \hline
        
            &  &  & \graycell{15.00} & \graycell{11.71} & \graycell{14.84} &  & \graycell{7.8} & \graycell{4.39} & \graycell{7.41} \\
        \multirow{-2}{*}{CLIP\textsuperscript{*}} & \multirow{-2}{*}{428M} & \multirow{-2}{*}{\graycell{10.64}} & +4.36 & +1.07 & +4.20 & \multirow{-2}{*}{\graycell{3.63}} & +4.17 & +0.76 & +3.78 \\

        \rowcolor{gray!10}
        
            &  &  & \graycell{21.01} & \graycell{22.29} & \graycell{21.81} &  & \graycell{14.01} & \graycell{12.44} & \graycell{12.52} \\
        \rowcolor{gray!10}
        \multirow{-2}{*}{dino.txt\textsuperscript{*}} & \multirow{-2}{*}{330M} & \multirow{-2}{*}{\graycell{19.81}} & +1.2 & +2.48 & +2.00 & \multirow{-2}{*}{\graycell{12.37}} & +1.64 & +0.07 & +0.15 \\

        RADIOv4 &  &  & \graycell{9.25} & \graycell{28.52} & \graycell{27.74} & & \graycell{4.79} & \graycell{28.8} & \graycell{28.89} \\
        \scriptsize{{SigLIP~2~Head}} & \multirow{-2}{*}{475M} & \multirow{-2}{*}{\graycell{25.93}} & -16.68 & +2.59 & +1.81 & \multirow{-2}{*}{\graycell{19.11}} & -14.32 & +9.69 & +9.78 \\

        \hline \hline
        \multicolumn{3}{l}{{Avg.~$\Delta$mIoU~$\uparrow$} (All)}
        & -4.23 & \underline{1.31} & \textbf{1.48} & & -1.88 & \underline{3.3} & \textbf{4.12} \\

        \multicolumn{3}{l}{{Avg.~$\Delta$mIoU~$\uparrow$} (Third-Party)}
        & -3.7 & \underline{1.56} & \textbf{1.83} & & +0.0 & \underline{2.9} & \textbf{4.05} \\ \thickhline

        \hline
    \end{tabular}
    \vspace{0.9mm}
    \caption{\textbf{2D OVSS.} mIoU and point-deltas across dense language-grounded backbones.
    Avg.~$\Delta$mIoU depicts average point-deltas through feature reconstruction.
    \textbf{Lite} and \textbf{Full} depict ReSiReg variants on top of backbones, while \textit{Calib} adopts the self-calibration baseline.
    \textit{BB} denotes backbone-only performance; reconstruction methods show absolute mIoU in grey and $\Delta$mIoU below in black.
    \textit{Param} depicts visual encoder parameters; \textsuperscript{*} marks lower input resolution.
    }
    \label{tab:2dovss}
    \vspace{-2mm}
\end{table}

For language-conditioned robot control tasks, perception models require simultaneously dense and language-grounded understanding from image- up to scene-level.
We evaluate quantitatively on OVSS and 3D semantic mapping, complemented by qualitative real-world language-conditioned manipulation.
Experiments span recent VLM backbones with and without our method.

\textbf{Baselines.}
For OVSS, ReSiReg is applied to CLIP \cite{Radford2021CLIP, Zhou2021MaskCLIP2}, DINOv2 \cite{Jose2025DinoTXT, Oquab2023DINOv2}, and RADIO \cite{Heinrich2024RADIO2}.
CLIP and DINOv2 have been established as seminal work for open-language and dense computer vision tasks respectively.
RADIO is trained through multi-teacher distillation, allowing multiple feature distributions to be recovered from a shared backbone.
RadSeg, built on RADIO with a modified SigLIP 2 \cite{Tschannen2025SigLIP2} head, is included as a backbone due to its strong empirical performance in general-purpose OVSS.
OTAS \cite{Schwaiger2025OTAS} and TIPSv2 \cite{Cao2026TIPSv2} are third party baselines due to their strong non-object-centric knowledge retrieval and explicit supervision of spatial consistency, respectively.
We adopt self-calibration from \cite{Bai2025SelfCalibratedCLIP} as a training-free feature reconstruction baseline. 
Due to the typically constrained on-board compute of robot systems, we further include a custom ViT-S-sized VLM trained on the EUPE \cite{Zhu2026EUPE} backbone through combined distillation from SigLIP 2 \cite{Tschannen2025SigLIP2} and contrastive global and patch-level conditioning \cite{Jose2025DinoTXT}.
Training protocol is provided in the appendix.

\textbf{Datasets and Evaluation Protocol.}
Evaluation follows \cite{alama2025radseg} at medium resolution of 576 on the shorter image side, using prompt templates applied over a set of possible target classes.
Backbones with a lower maximum resolution are run at their maximum resolution with bilinear upscaling of output features.
In order to minimise potential masking of prediction noise, we omit prompt denoising, sliding window inference, and mask refinement networks.
Patch-level feature maps are upscaled using nearest neighbour interpolation.
Patch-wise 3D feature aggregation follows \cite{Schwaiger2025OTAS}, similarly to 2D we omit neighbour-based prediction smoothing.
We report mean intersection over union (mIoU), frequency-weighted mIoU (f-mIoU) in 3D, and mIoU delta gained by the feature reconstruction methods.
Evaluation is done on ADE20K \cite{ADE20k} for general-purpose OVSS, ORAD-3D \cite{Min2025ORAD3D} for unstructured outdoor domains, and ScanNet \cite{Dai2017Scannet} for 3D feature aggregation.

\textbf{Implementation Detail.}
We provide two implementations of our method with different accuracy-latency trade-offs.
\textit{ReSiReg Full} fully implements our method, using Principal Component Analysis (PCA) as the LVM $\phi$ and K-means as the clustering model $\mathcal{C}$ with hyperparameters $r=n=36$ and softmax temperature $\alpha=5.0$.
Choice of $\phi$ and $\mathcal{C}$ follows \cite{Schwaiger2025OTAS}.
\textit{ReSiReg Lite} approximates clustering by building a self-similarity matrix \cite{alama2025radseg, Shi2025Trident, Lan2025ProxyCLIP} from model intermediates using $Simmat = F_v^{flat} \cdot {F_v^{flat}}^{T}$.
Temperature-scaled softmax \cite{Hinton2015Distilling} with $\alpha=10.$ is applied column-wise over $Simmat$ and masked to obtain affinities.
This enables subsequent language-grounded feature aggregation, while dropping per-batch test-time optimisation of $\phi$ and $\mathcal{C}$.
Both use $\tau=0$.

\begin{figure}
    \centering
    \includegraphics[width=0.99\linewidth]{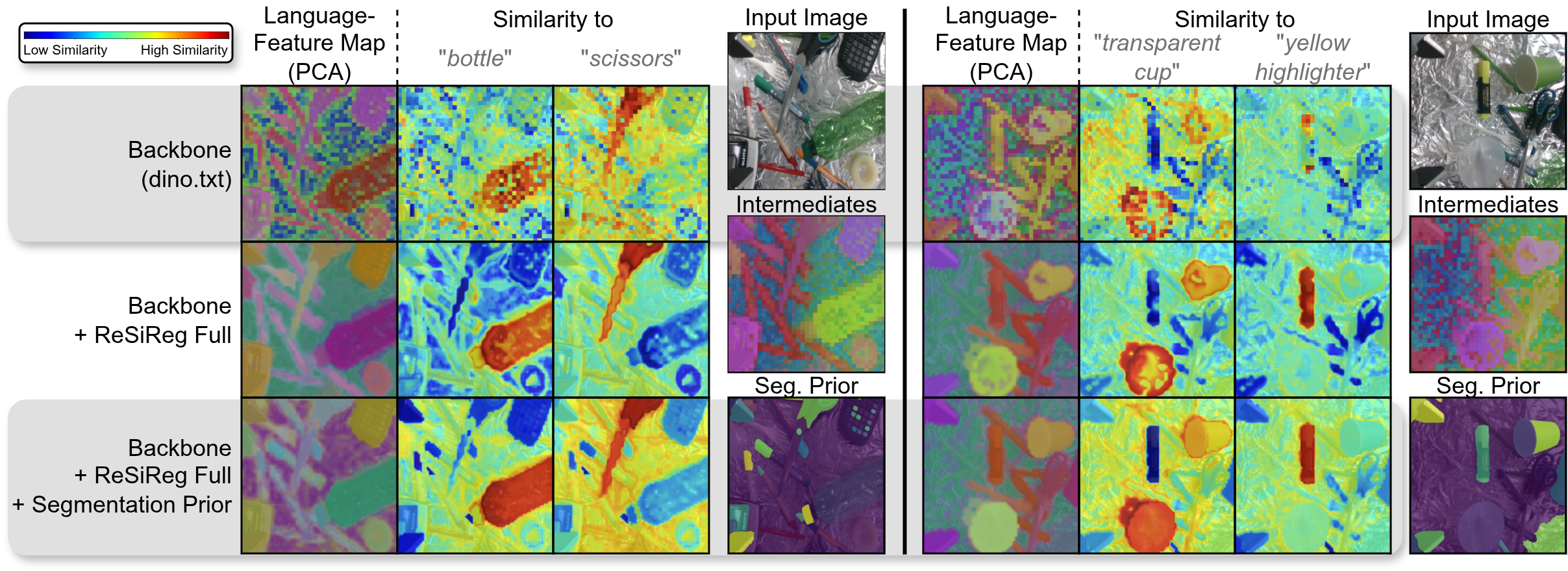}
    \vspace{-3mm}
    \caption{\textbf{Robotic manipulation stress test.} Dense similarity maps in a cluttered grasping scene with reflective, transparent, and overlapping objects. ReSiReg improves spatial consistency over the backbone output, while the optional segmentation prior sharpens object boundaries when available.}
    \label{fig:qual_manipulation}
    \vspace{-2mm}
\end{figure}

\subsection{Open-Vocabulary Semantic Segmentation}

Tab.~\ref{tab:2dovss} presents results on 2D OVSS.
Both ReSiReg variants produce varying levels of uplift across third party backbones.
In contrast, self-calibration depends on the used model and with improvements on CLIP and dino.txt, while significantly decreasing mIoU for RADIO on ADE20K.
Improvement on our 25M VLM is more domain dependent, showing a small performance decrease on ADE20K, with significant improvement on ORAD-3D.
Due to the low parameter count, the model does not fully capture abstract language semantics.
Therefore, ReSiReg's mixtures may dilute correct mask assignment.
To separate this effect from third-party backbones, we also report average deltas excluding our VLM.
Interestingly, a majority of backbones show larger performance improvement on ORAD-3D.
This indicates a weaker language-feature structure due to the lower representation of the outdoor domain in annotated training data.
RadSeg already performs similarity matrix aggregation in its attention mechanism and output, compounding diminishing deltas on ADE20K.
However, both ReSiReg methods result in IoU improvements for RadSeg on ORAD-3D.  
ViT-L results follow prior work highlighting improved dense accuracy of smaller ViT-B backbones due to distillation \cite{Cao2026TIPSv2}.
Especially for RADIOv4 on ORAD-3D, ReSiReg closes this accuracy gap.


\subsection{Scene-Level 3D Aggregation}

Tab.~\ref{tab:3dscannet} presents 3D feature aggregation on ScanNet.
To emphasise raw encoder performance, patch-level per-frame features are fused through 3D projection and voxel-downsampling. Evaluation
\begin{wraptable}[9]{r}{7.2cm}
    \begin{tabular}{l"c c c}
         \thickhline
         \textbf{Model} & \textbf{Param}$\downarrow$ & \textbf{mIoU}$\uparrow$ & \textbf{f-mIoU}$\uparrow$ \\
         
         TIPSv2 & 86M & 21.91 & 22.97 \\
         OTAS & 105M & 28.05 & 37.18 \\ \hline

        EUPE+Head\textsuperscript{*} & 25M & 18.59 & 19.91 \\
        CLIP\textsuperscript{*} & 86M & 20.84 & 29.96 \\
        \thickhline
    
    \end{tabular}
    \caption{\textbf{ScanNet Feature Aggregation}. Segmentation accuracy of 3D-projected VLM features. \textsuperscript{*}~indicates models with ReSiReg Full.}
    \label{tab:3dscannet}
\end{wraptable}
includes the EUPE-based 25M-parameter backbone, CLIP, both with ReSiReg Full against third party baselines.
Our ViT-S backbone and CLIP both approach TIPSv2.
ReSiReg therefore narrows the gap between model sizes for EUPE and training objective for CLIP, since TIPSv2 is trained on a joint embedding objective in addition to a contrastive loss.


\subsection{Open-Vocabulary Robotic Manipulation}

To demonstrate ReSiReg in the real-world, we apply it to a heavily cluttered robotic manipulation scene.
The scene features reflective foil background, overlapping, and transparent objects as a stress-test for the VLM encoder.
Fig.~\ref{fig:qual_manipulation} compares dense similarity maps from the backbone, ReSiReg Full, and ReSiReg Full with SAM2.1 Hiera-T \cite{ravi2024sam2} as the optional segmentation prior.
The backbone produces noisy and fragmented activations, while ReSiReg improves spatial consistency of the queried target regions.
The optional segmentation prior further sharpens object boundaries.


\subsection{Ablations} 

\begin{figure}
    \centering
    \includegraphics[width=0.99\linewidth]{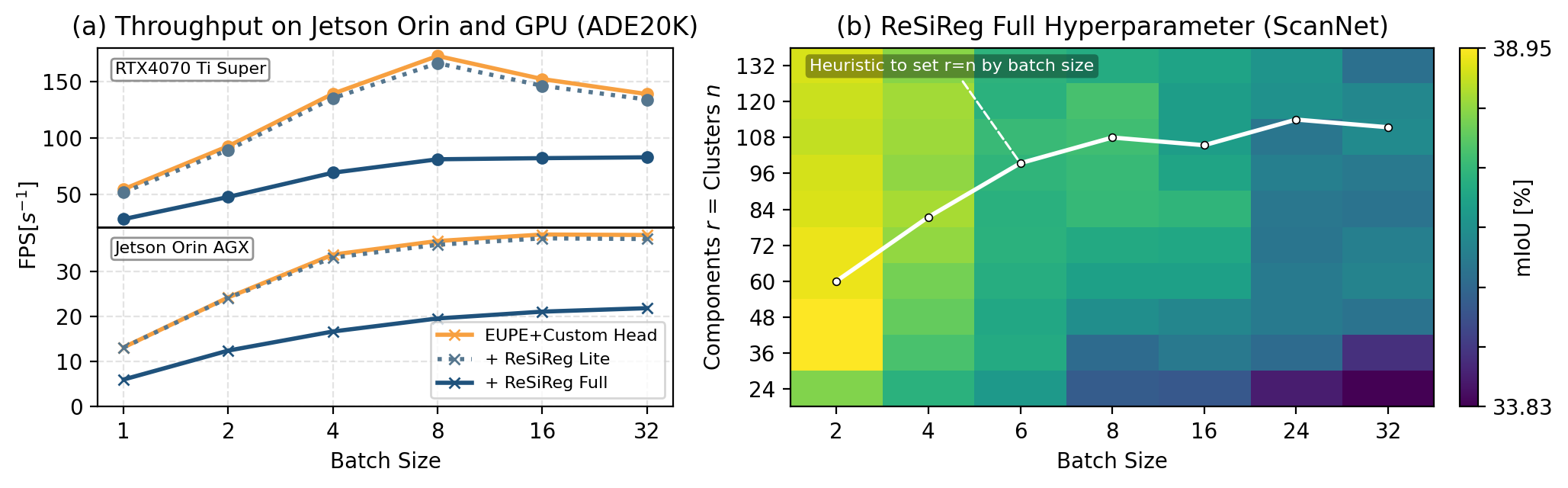}
    \vspace{-4mm}
    \caption{\textbf{Deployment and prototype selection ablations.} Left: Runtime of our 25M VLM, ReSiReg Lite, and ReSiReg Full on Jetson and GPU. Right: ScanNet 3D aggregation mIoU for ReSiReg Full on RadSeg over tied cluster/component counts with resulting hyperparameter heuristic.}
    \label{fig:ablations}
    \vspace{-2mm}
\end{figure}

\textbf{Runtime on embedded hardware.}
To evaluate applicability under robotic onboard compute constraints, we evaluate the runtime of the 25M EUPE-based dense VLM.
Fig.~\ref{fig:ablations}a) plots effective frames per second (fps) on GPU and NVIDIA Jetson Orin, comparing the backbone, +ReSiReg Lite, and +ReSiReg Full across batch sizes.
On Jetson, ReSiReg Lite and backbone-only surpass 30fps in batched inference.
ReSiReg Full is more expensive with roughly 30\% less throughput.

\textbf{Batch-size and prototype heuristic.}
ReSiReg Full optimises the LVM and clustering model over the input batch.
We therefore evaluate the interaction between batch size and the number of visual prototypes on ScanNet 3D aggregation over the RadSeg backbone (see Fig.~\ref{fig:ablations}b).
Following \cite{Schwaiger2025OTAS}, we tie the number of clusters and LVM components and report mIoU as a heatmap.
The resulting trend is used to derive a heuristic for selecting the number of prototypes from the batch size.


\section{Limitations}
\label{sec:limitations}
ReSiReg improves spatial feature consistency rather than class-level language alignment.
Thus, dense retrieval is improved only when the target concept is already represented by the underlying VLM, rather than recovering missing or weak language alignment.
This is reflected in challenging many-class segmentation settings such as ADE20K, where improvements are more limited and can even become negative for weaker dense grounding.
In such cases, prototype mixing may smooth spatial predictions but dilute class-specific evidence.
ReSiReg assumes spatially consistent VLM intermediates, which is typically the case for distillation-based backbones or language heads trained on foundation models.
In contrast, \cite{Cao2026TIPSv2} jointly optimise SSL-like and contrastive language grounding; future work should investigate whether such models benefit from post-hoc reconstruction.

ReSiReg Full optimises the LVM and clustering model over the input batch.
This improves efficiency at larger batch sizes, but assumes jointly processed images depict the same scene or a temporally coherent sequence.
This matches robotic settings such as semantic mapping and OVSS on video streams, but may not hold for unrelated image batches.
Further, applying ReSiReg Full during VLM training is limited by the test-time optimised LVM and clustering, whose fitting and hard assignment steps do not provide stable gradient flow.
ReSiReg Lite avoids this through approximating clustering with self-similarity, but trades the explicit prototype structure of ReSiReg Full.

Our choice of LVM and clustering methods follows \cite{Schwaiger2025OTAS}.
Compared to probabilistic clustering methods, this accelerates computation, especially on embedded robotics hardware (see Fig.~\ref{fig:ablations}), but requires empirical tuning of the number of prototypes.
Future work should therefore investigate alternatives, such as optimising the number of prototypes using the evidence lower bound for variational Bayesian Gaussian mixture models.


\section{Conclusion}
\label{sec:conclusion}
We introduced ReSiReg, a feature reconstruction method that uses spatially consistent VLM intermediate features to reconstruct dense language embeddings as soft mixtures of visual prototypes.
Across open-vocabulary semantic segmentation, 3D semantic mapping, and qualitative language-conditioned manipulation, ReSiReg improved dense semantic retrieval without increasing the number of model parameters.
Results indicate that spatial structure of VLM intermediates can improve dense language-grounded representations for backbones and language heads with already strong language alignment.
Based on these findings, we provide a compact ViT-S VLM for robotic tasks.


\clearpage

\acknowledgments{This work was partly supported by the city of Vienna (MA23 – Economic Affairs, Labour and Statistics) through the project Stadt Wien Kompetenzteam für Drohnentechnik in der Fachhochschulausbildung (DrohnFH, MA23 project 35-02).}


\bibliography{example}  

@inproceedings{Ansel2024PyTorch,
author = {Ansel, Jason and Yang, Edward and He, Horace and Gimelshein, Natalia and Jain, Animesh and Voznesensky, Michael and Bao, Bin and Bell, Peter and Berard, David and Burovski, Evgeni and Chauhan, Geeta and Chourdia, Anjali and Constable, Will and Desmaison, Alban and DeVito, Zachary and Ellison, Elias and Feng, Will and Gong, Jiong and Gschwind, Michael and Hirsh, Brian and Huang, Sherlock and Kalambarkar, Kshiteej and Kirsch, Laurent and Lazos, Michael and Lezcano, Mario and Liang, Yanbo and Liang, Jason and Lu, Yinghai and Luk, CK and Maher, Bert and Pan, Yunjie and Puhrsch, Christian and Reso, Matthias and Saroufim, Mark and Siraichi, Marcos Yukio and Suk, Helen and Suo, Michael and Tillet, Phil and Wang, Eikan and Wang, Xiaodong and Wen, William and Zhang, Shunting and Zhao, Xu and Zhou, Keren and Zou, Richard and Mathews, Ajit and Chanan, Gregory and Wu, Peng and Chintala, Soumith},
booktitle = {29th ACM International Conference on Architectural Support for Programming Languages and Operating Systems, Volume 2 (ASPLOS '24)},
doi = {10.1145/3620665.3640366},
month = apr,
publisher = {ACM},
title = {{PyTorch 2: Faster Machine Learning Through Dynamic Python Bytecode Transformation and Graph Compilation}},
url = {https://docs.pytorch.org/assets/pytorch2-2.pdf},
year = {2024}
}

@article{Oquab2023DINOv2,
  title   = {DINOv2: Learning Robust Visual Features without Supervision},
  author  = {Oquab, Maxime and Darcet, Timoth{\'e}e and Moutakanni, Th{\'e}o and Vo, Huy V. and Szafraniec, Marc and Khalidov, Vasil and Fernandez, Pierre and Haziza, Daniel and Massa, Francisco and El-Nouby, Alaaeldin and Howes, Russell and Huang, Po-Yao and Xu, Hu and Sharma, Vasu and Li, Shang-Wen and Galuba, Wojciech and Rabbat, Mike and Assran, Mido and Ballas, Nicolas and Synnaeve, Gabriel and Misra, Ishan and J{\'e}gou, Herv{\'e} and Mairal, Julien and Labatut, Patrick and Joulin, Armand and Bojanowski, Piotr},
  year    = {2023},
  eprint  = {2304.07193},
  archivePrefix = {arXiv},
  primaryClass  = {cs.CV},
  doi     = {10.48550/arXiv.2304.07193},
  url     = {https://arxiv.org/abs/2304.07193}
}

@article{Darcet2023VitRegisters,
  title   = {Vision Transformers Need Registers},
  author  = {Darcet, Timoth{\'e}e and Oquab, Maxime and Mairal, Julien and Bojanowski, Piotr},
  year    = {2023},
  eprint  = {2309.16588},
  archivePrefix = {arXiv},
  primaryClass  = {cs.CV},
  doi     = {10.48550/arXiv.2309.16588},
  url     = {https://arxiv.org/abs/2309.16588}
}

@article{Simeoni2025Dinov3,
  title   = {DINOv3},
  author  = {Sim{\'e}oni, Oriane and Vo, Huy V. and Seitzer, Maximilian and Baldassarre, Federico and Oquab, Maxime and Jose, Cijo and Khalidov, Vasil and Szafraniec, Marc and Yi, Seungeun and Ramamonjisoa, Micha{\"e}l and Massa, Francisco and Haziza, Daniel and Wehrstedt, Luca and Wang, Jianyuan and Darcet, Timoth{\'e}e and Moutakanni, Th{\'e}o and Sentana, Leonel and Roberts, Claire and Vedaldi, Andrea and Tolan, Jamie and Brandt, John and Couprie, Camille and Mairal, Julien and J{\'e}gou, Herv{\'e} and Labatut, Patrick and Bojanowski, Piotr},
  year    = {2025},
  eprint  = {2508.10104},
  archivePrefix = {arXiv},
  primaryClass  = {cs.CV},
  doi     = {10.48550/arXiv.2508.10104},
  url     = {https://arxiv.org/abs/2508.10104}
}

@InProceedings{Ranzinger2024AMRADIO,
    author    = {Ranzinger, Mike and Heinrich, Greg and Kautz, Jan and Molchanov, Pavlo},
    title     = {AM-RADIO: Agglomerative Vision Foundation Model Reduce All Domains Into One},
    booktitle = {Proceedings of the IEEE/CVF Conference on Computer Vision and Pattern Recognition (CVPR)},
    month     = {June},
    year      = {2024},
    pages     = {12490-12500}
}

@InProceedings{Heinrich2024RADIO2,
    author    = {Heinrich, Greg and Ranzinger, Mike and Yin, Hongxu and Lu, Yao and Kautz, Jan and Tao, Andrew and Catanzaro, Bryan and Molchanov, Pavlo},
    title     = {RADIOv2.5: Improved Baselines for Agglomerative Vision Foundation Models},
    booktitle = {Proceedings of the IEEE/CVF Conference on Computer Vision and Pattern Recognition (CVPR)},
    year      = {2025},
    pages     = {22487-22497}
}

@InProceedings{Cherti2023OpenCLIP,
    author    = {Cherti, Mehdi and Beaumont, Romain and Wightman, Ross and Wortsman, Mitchell and Ilharco, Gabriel and Gordon, Cade and Schuhmann, Christoph and Schmidt, Ludwig and Jitsev, Jenia},
    title     = {Reproducible Scaling Laws for Contrastive Language-Image Learning},
    booktitle = {Proceedings of the IEEE/CVF Conference on Computer Vision and Pattern Recognition (CVPR)},
    year      = {2023},
    pages     = {2818-2829}
}

@InProceedings{Radford2021CLIP,
  title = 	 {Learning Transferable Visual Models From Natural Language Supervision},
  author =       {Radford, Alec and Kim, Jong Wook and Hallacy, Chris and Ramesh, Aditya and Goh, Gabriel and Agarwal, Sandhini and Sastry, Girish and Askell, Amanda and Mishkin, Pamela and Clark, Jack and Krueger, Gretchen and Sutskever, Ilya},
  booktitle = 	 {Proceedings of the 38th International Conference on Machine Learning},
  pages = 	 {8748--8763},
  year = 	 {2021},
  editor = 	 {Meila, Marina and Zhang, Tong},
  volume = 	 {139},
  series = 	 {Proceedings of Machine Learning Research},
  publisher =    {PMLR}
}

@InProceedings{Zhou2021MaskCLIP2,
author={Zhou, Chong and Loy, Chen Change and Dai, Bo},
editor={Avidan, Shai and Brostow, Gabriel and Ciss{\'e}, Moustapha and Farinella, Giovanni Maria and Hassner, Tal},
title={Extract Free Dense Labels from CLIP},
booktitle={Computer Vision -- ECCV 2022},
year={2022},
publisher={Springer Nature Switzerland},
address={Cham},
pages={696-712}
}

@article{alama2025radseg,
  title   = {RADSeg: Unleashing Parameter and Compute Efficient Zero-Shot Open-Vocabulary Segmentation Using Agglomerative Models},
  author  = {Alama, Omar and Jariwala, Darshil and Bhattacharya, Avigyan and Kim, Seungchan and Wang, Wenshan and Scherer, Sebastian},
  year    = {2025},
  eprint  = {2511.19704},
  archivePrefix = {arXiv},
  primaryClass  = {cs.CV},
  doi     = {10.48550/arXiv.2511.19704},
  url     = {https://arxiv.org/abs/2511.19704}
}

@article{Cao2026TIPSv2,
  title   = {TIPSv2: Advancing Vision-Language Pretraining with Enhanced Patch-Text Alignment},
  author  = {Cao, Bingyi and Chen, Koert and Maninis, Kevis-Kokitsi and Chen, Kaifeng and Karpur, Arjun and Xia, Ye and Dua, Sahil and Dabral, Tanmaya and Han, Guangxing and Han, Bohyung and Ainslie, Joshua and Bewley, Alex and Jacob, Mithun and Wagner, Ren{\'e} and Ramos, Washington and Choromanski, Krzysztof and Seyedhosseini, Mojtaba and Zhou, Howard and Araujo, Andr{\'e}},
  year    = {2026},
  eprint  = {2604.12012},
  archivePrefix = {arXiv},
  primaryClass  = {cs.CV},
  doi     = {10.48550/arXiv.2604.12012},
  url     = {https://arxiv.org/abs/2604.12012}
}

@article{Bolya2025PerceptionEncoder,
  title   = {Perception Encoder: The best visual embeddings are not at the output of the network},
  author  = {Bolya, Daniel and Huang, Po-Yao and Sun, Peize and Cho, Jang Hyun and Madotto, Andrea and Wei, Chen and Ma, Tengyu and Zhi, Jiale and Rajasegaran, Jathushan and Rasheed, Hanoona and Wang, Junke and Monteiro, Marco and Xu, Hu and Dong, Shiyu and Ravi, Nikhila and Li, Daniel and Doll{\'a}r, Piotr and Feichtenhofer, Christoph},
  year    = {2025},
  eprint  = {2504.13181},
  archivePrefix = {arXiv},
  primaryClass  = {cs.CV},
  doi     = {10.48550/arXiv.2504.13181},
  url     = {https://arxiv.org/abs/2504.13181}
}

@article{Tschannen2025SigLIP2,
  title   = {SigLIP 2: Multilingual Vision-Language Encoders with Improved Semantic Understanding, Localization, and Dense Features},
  author  = {Tschannen, Michael and Gritsenko, Alexey and Wang, Xiao and Naeem, Muhammad Ferjad and Alabdulmohsin, Ibrahim and Parthasarathy, Nikhil and Evans, Talfan and Beyer, Lucas and Xia, Ye and Mustafa, Basil and H{\'e}naff, Olivier and Harmsen, Jeremiah and Steiner, Andreas and Zhai, Xiaohua},
  year    = {2025},
  eprint  = {2502.14786},
  archivePrefix = {arXiv},
  primaryClass  = {cs.CV},
  doi     = {10.48550/arXiv.2502.14786},
  url     = {https://arxiv.org/abs/2502.14786}
}

@article{Zhu2026EUPE,
  title   = {Efficient Universal Perception Encoder},
  author  = {Zhu, Chenchen and Suri, Saksham and Jose, Cijo and Oquab, Maxime and Szafraniec, Marc and Wen, Wei and Xiong, Yunyang and Labatut, Patrick and Bojanowski, Piotr and Krishnamoorthi, Raghuraman and Chandra, Vikas},
  year    = {2026},
  eprint  = {2603.22387},
  archivePrefix = {arXiv},
  primaryClass  = {cs.CV},
  doi     = {10.48550/arXiv.2603.22387},
  url     = {https://arxiv.org/abs/2603.22387}
}

@article{ravi2024sam2,
  title={SAM 2: Segment Anything in Images and Videos},
  author={Ravi, Nikhila and Gabeur, Valentin and Hu, Yuan-Ting and Hu, Ronghang and Ryali, Chaitanya and Ma, Tengyu and Khedr, Haitham and R{\"a}dle, Roman and Rolland, Chloe and Gustafson, Laura and Mintun, Eric and Pan, Junting and Alwala, Kalyan Vasudev and Carion, Nicolas and Wu, Chao-Yuan and Girshick, Ross and Doll{\'a}r, Piotr and Feichtenhofer, Christoph},
  journal={arXiv preprint arXiv:2408.00714},
  url={https://arxiv.org/abs/2408.00714},
  year={2024}
}

@ARTICLE{Bai2025SelfCalibratedCLIP,
  author={Bai, Sule and Liu, Yong and Han, Yifei and Zhang, Haoji and Tang, Yansong and Zhou, Jie and Lu, Jiwen},
  journal={IEEE Transactions on Image Processing}, 
  title={Self-Calibrated CLIP for Training-Free Open-Vocabulary Segmentation}, 
  year={2025},
  volume={34},
  number={},
  pages={8271-8284}}

@InProceedings{Wysoczanska2025CLIPDinoiser,
author={Wysocza{\'{n}}ska, Monika
and Sim{\'e}oni, Oriane
and Ramamonjisoa, Micha{\"e}l
and Bursuc, Andrei
and Trzci{\'{n}}ski, Tomasz
and P{\'e}rez, Patrick},
editor={Leonardis, Ale{\v{s}}
and Ricci, Elisa
and Roth, Stefan
and Russakovsky, Olga
and Sattler, Torsten
and Varol, G{\"u}l},
title={CLIP-DINOiser: Teaching CLIP a Few DINO Tricks for Open-Vocabulary Semantic Segmentation},
booktitle={Computer Vision -- ECCV 2024},
year={2025},
publisher={Springer Nature Switzerland},
address={Cham},
pages={320-337}
}

@InProceedings{Wang2025SClip,
author={Wang, Feng
and Mei, Jieru
and Yuille, Alan},
editor={Leonardis, Ale{\v{s}}
and Ricci, Elisa
and Roth, Stefan
and Russakovsky, Olga
and Sattler, Torsten
and Varol, G{\"u}l},
title={SCLIP: Rethinking Self-Attention for Dense Vision-Language Inference},
booktitle={Computer Vision -- ECCV 2024},
year={2025},
publisher={Springer Nature Switzerland},
address={Cham},
pages={315-332}
}

@InProceedings{Jose2025DinoTXT,
    author    = {Jose, Cijo and Moutakanni, Th\'eo and Kang, Dahyun and Baldassarre, Federico and Darcet, Timoth\'ee and Xu, Hu and Li, Daniel and Szafraniec, Marc and Ramamonjisoa, Micha\"el and Oquab, Maxime and Sim\'eoni, Oriane and Vo, Huy V. and Labatut, Patrick and Bojanowski, Piotr},
    title     = {DINOv2 Meets Text: A Unified Framework for Image- and Pixel-Level Vision-Language Alignment},
    booktitle = {Proceedings of the IEEE/CVF Conference on Computer Vision and Pattern Recognition (CVPR)},
    month     = {June},
    year      = {2025},
    pages     = {24905-24916}
}

@InProceedings{Lan2025ProxyCLIP,
author={Lan, Mengcheng
and Chen, Chaofeng
and Ke, Yiping
and Wang, Xinjiang
and Feng, Litong
and Zhang, Wayne},
editor={Leonardis, Ale{\v{s}}
and Ricci, Elisa
and Roth, Stefan
and Russakovsky, Olga
and Sattler, Torsten
and Varol, G{\"u}l},
title={ProxyCLIP: Proxy Attention Improves CLIP for Open-Vocabulary Segmentation},
booktitle={Computer Vision -- ECCV 2024},
year={2025},
publisher={Springer Nature Switzerland},
address={Cham},
pages={70-88}}

@InProceedings{Shi2025Trident,
    author    = {Shi, Yuheng and Dong, Minjing and Xu, Chang},
    title     = {Harnessing Vision Foundation Models for High-Performance, Training-Free Open Vocabulary Segmentation},
    booktitle = {Proceedings of the IEEE/CVF International Conference on Computer Vision (ICCV)},
    month     = {October},
    year      = {2025},
    pages     = {23487-23497}
}

@article{Schwaiger2025OTAS,
  title   = {OTAS: Open-vocabulary Token Alignment for Outdoor Segmentation},
  author  = {Schwaiger, Simon and Thalhammer, Stefan and W{\"o}ber, Wilfried and Steinbauer-Wagner, Gerald},
  year    = {2025},
  eprint  = {2507.08851},
  archivePrefix = {arXiv},
  primaryClass  = {cs.CV},
  doi     = {10.48550/arXiv.2507.08851},
  url     = {https://arxiv.org/abs/2507.08851}
}

@INPROCEEDINGS{Alama2025Rayfronts,
  author={Alama, Omar and Bhattacharya, Avigyan and He, Haoyang and Kim, Seungchan and Qiu, Yuheng and Wang, Wenshan and Ho, Cherie and Keetha, Nikhil and Scherer, Sebastian},
  booktitle={2025 IEEE/RSJ International Conference on Intelligent Robots and Systems (IROS)}, 
  title={RayFronts: Open-Set Semantic Ray Frontiers for Online Scene Understanding and Exploration}, 
  year={2025},
  volume={},
  number={},
  pages={5930-5937}
}

@InProceedings{Hajimiri2025NaClip,
    author    = {Hajimiri, Sina and Ben Ayed, Ismail and Dolz, Jose},
    title     = {Pay Attention to Your Neighbours: Training-Free Open-Vocabulary Semantic Segmentation},
    booktitle = {Proceedings of the Winter Conference on Applications of Computer Vision (WACV)},
    month     = {February},
    year      = {2025},
    pages     = {5061-5071}
}

@article{Hinton2015Distilling,
  title         = {Distilling the Knowledge in a Neural Network},
  author        = {Hinton, Geoffrey and Vinyals, Oriol and Dean, Jeff},
  journal       = {arXiv preprint arXiv:1503.02531},
  year          = {2015},
  eprint        = {1503.02531},
  archivePrefix = {arXiv},
  primaryClass  = {stat.ML},
  doi           = {10.48550/arXiv.1503.02531}
}

@inproceedings{Qiu2024Featuresplatting,
      title={Language-Driven Physics-Based Scene Synthesis and Editing via Feature Splatting},
      author={Ri-Zhao Qiu and Ge Yang and Weijia Zeng and Xiaolong Wang},
      booktitle={European Conference on Computer Vision (ECCV)},
      year={2024},
      pages={368-383}
    }

@inproceedings{Yamazaki2024OpenFusion,
  title={Open-fusion: Real-time open-vocabulary 3d mapping and queryable scene representation},
  author={Yamazaki, Kashu and Hanyu, Taisei and Vo, Khoa and Pham, Thang and Tran, Minh and Doretto, Gianfranco and Nguyen, Anh and Le, Ngan},
  booktitle={2024 IEEE International Conference on Robotics and Automation (ICRA)},
  pages={9411--9417},
  year={2024},
}

@INPROCEEDINGS{Kerr2023LERF,
    author              = {Kerr, Justin and Kim, Chung Min and Goldberg, Ken and Kanazawa, Angjoo and Tancik, Matthew},
    booktitle           = {2023 IEEE/CVF International Conference on Computer Vision (ICCV)}, 
    title               = {LERF: Language Embedded Radiance Fields}, 
    year                = {2023},
    volume              = {},
    number              = {},
    pages               = {19672-19682},
}

@phdthesis{woeber2023nonlinear,
  author  = {W{\"o}ber, Wilfried},
  title   = {Nonlinear and nonparametric methods for statistical shape analysis},
  school  = {University of Natural Resources and Life Sciences, Vienna (BOKU)},
  address = {Vienna, Austria},
  year    = {2023},
  type    = {Doctoral dissertation},
  url     = {https://epub.boku.ac.at/obvbokhs/content/titleinfo/11864305}
}

@inproceedings{Zhou2022iBOT,
  title     = {Image BERT Pre-training with Online Tokenizer},
  author    = {Zhou, Jinghao and Wei, Chen and Wang, Huiyu and Shen, Wei and Xie, Cihang and Yuille, Alan and Kong, Tao},
  booktitle = {International Conference on Learning Representations (ICLR)},
  year      = {2022}
}

@inproceedings{man2024lexicon3d,
      title={Lexicon3D: Probing Visual Foundation Models for Complex 3D Scene Understanding},
      author={Man, Yunze and Zheng, Shuhong and Bao, Zhipeng and Hebert, Martial and Gui, Liang-Yan and Wang, Yu-Xiong},
      booktitle={Advances in Neural Information Processing Systems},
      year={2024} 
      }

@inproceedings{huang2023voxposer,
  title={VoxPoser: Composable 3D Value Maps for Robotic Manipulation with Language Models},
  author={Huang, Wenlong and Wang, Chen and Zhang, Ruohan and Li, Yunzhu and Wu, Jiajun and Fei-Fei, Li},
  booktitle={Conference on Robot Learning},
  pages={540--562},
  year={2023},
  organization={PMLR}
}

@Article{ADE20k,
author={Zhou, Bolei
and Zhao, Hang
and Puig, Xavier
and Xiao, Tete
and Fidler, Sanja
and Barriuso, Adela
and Torralba, Antonio},
title={Semantic Understanding of Scenes Through the ADE20K Dataset},
journal={International Journal of Computer Vision},
year={2019},
month={Mar},
day={01},
volume={127},
number={3},
pages={302-321},
issn={1573-1405},
doi={10.1007/s11263-018-1140-0},
}

@article{Min2025ORAD3D,
  title   = {Advancing Off-Road Autonomous Driving: The Large-Scale ORAD-3D Dataset and Comprehensive Benchmarks},
  author  = {Min, Chen and Mei, Jilin and Zhai, Heng and Wang, Shuai and Sun, Tong and Kong, Fanjie and Li, Haoyang and Mao, Fangyuan and Liu, Fuyang and Wang, Shuo and Nie, Yiming and Zhu, Qi and Xiao, Liang and Zhao, Dawei and Hu, Yu},
  year    = {2025},
  eprint  = {2510.16500},
  archivePrefix = {arXiv},
  primaryClass  = {cs.RO},
  doi     = {10.48550/arXiv.2510.16500},
  url     = {https://arxiv.org/abs/2510.16500}
}

@InProceedings{Dai2017Scannet,
author = {Dai, Angela and Chang, Angel X. and Savva, Manolis and Halber, Maciej and Funkhouser, Thomas and Niessner, Matthias},
title = {ScanNet: Richly-Annotated 3D Reconstructions of Indoor Scenes},
booktitle = {Proceedings of the IEEE Conference on Computer Vision and Pattern Recognition (CVPR)},
month = {July},
year = {2017}
}

\clearpage
\appendix

\begin{center}
  \large\bfseries Supplementary Material for \textbf{ReSiReg: Towards Spatially Consistent Semantics in Language-Conditioned Robotic Tasks}
\end{center}

\section{Experimental Setup}

\textbf{Input Resolutions.}
2D evaluations resize inputs with a shorter side of 576 pixels while preserving aspect ratio.
EUPE, RADIO, and RadSeg are run at this resolution.
TIPSv2 uses a patch size of 14 and is therefore evaluated at 574 pixels on the shorter side.
OTAS and dino.txt are limited by the DINOv2 positional-encoding grid and are evaluated at a fixed $518\times518$ input.
CLIP baselines use each checkpoint's native input resolution ($224$ for ViT-B/16 and $336$ for ViT-L/14@336px), and dense outputs are bilinearly upsampled to the evaluation canvas where needed.
This applies to OTAS' CLIP path as well.

\textbf{Inference Settings.}
To isolate potential numerical instabilities, evaluations disable mixed precision and execute in FP32 on CUDA.
ReSiReg upscales intermediate backbone features by a factor of $s{=}2$ before clustering and reconstruction.

\textbf{Prompt Composition.}
Dense features are scaled with cosine similarity to text prototypes built from the OpenAI ImageNet template bank (80 templates per class).
For each class name $c$, every template prompt is embedded with the backbone text encoder, $\ell_2$-normalised embeddings are averaged, and re-normalised to obtain one query per class.
Per-pixel predictions use a temperature-scaled softmax over class similarities with temperature $100$.

ADE20K uses prompts over the fixed 150-class vocabulary from RADSeg \cite{alama2025radseg}, matching the ADE150 OVSS benchmark.
On ORAD-3D, annotations are sparse polygon labels with a small set of annotated target concepts depicting scene-structure.
Since no established vocabulary exists for this benchmark, we evaluate over the annotated classes: \textit{road, traversable ground, car, person, water, snow, grass on road, rock, sky,} and \textit{background}.
Polygon tags from the dataset are mapped to these names.
ScanNet 3D uses the same template-averaging protocol, with class names taken from the ScanNet\,v2 NYU40 label map and excluding \textit{otherprop}, \textit{otherstructure}, and \textit{otherfurniture}, following prior work.

\textbf{ScanNet 3D Aggregation.}
The 3D feature aggregation experiment evaluates scene-level semantic retrieval on twelve ScanNet\,v2 validation scenes, following \cite{alama2025radseg, Alama2025Rayfronts}.
Frames are subsampled with a frame skip of $10$; invalid camera poses are dropped.
RGB and depth images are resized to $640{\times}864$, with intrinsics scaled accordingly.
Per frame, dense language-grounded patch features are extracted and projected to 3D following \cite{Schwaiger2025OTAS}.
Patch centres are unprojected with the depth camera model; depth is aggregated per patch by nearest downsampling to feature resolution and back to the depth grid.
Points are transformed and fused across frames using voxel downsampling at $0.05$\,m, averaging features within each voxel and $\ell_2$-normalising the result.
Voxel labels are assigned by cosine similarity to the text prototypes, following the 2D OVSS experiments.
Ground-truth voxels are built independently from filtered 2D semantic labels, depth, and pose using the same voxel size and label mapping.
Predictions are transferred to ground truth voxel centres by nearest neighbour without spatial smoothing.

\textbf{Model Throughput Ablation Protocol.}
Runtime is measured on ADE20K validation with our 25M EUPE-based backbone, ReSiReg Lite, and ReSiReg Full, at $576$-px shorter-side resolution over batch sizes $\{1,2,4,8,16,32\}$.
Timing includes image embedding only (dense language-grounded encoding), excluding downstream text encoding and metrics computation.
Throughput experiments enable mixed precision using CUDA with bfloat16.

\textbf{Prototype Selection Ablation Protocol.}
The batch-size versus prototype-count ablation is a full factorial over ReSiReg Full on RadSeg with tied number of components and clusters.
Batch sizes are $\{2,4,6,8,16,24,32\}$ and prototype counts $\{24,36,48,60,72,84,96,108,120,132\}$.
To constrain computational requirements due to the large number of parameter combinations, evaluation parameters are adjusted compared to the main feature aggregation experiments.
The ablation uses frame skip of $20$ and voxel size $0.2$m with reduced ground truth depth resolution of $228{\times}308$.

\section{EUPE Language Head Training}

Our 25M parameter VLM uses a frozen EUPE ViT-S backbone \cite{Zhu2026EUPE} and trains a language head to align dense image features with SigLIP~2 \cite{Tschannen2025SigLIP2} text embeddings.
Training uses image-caption pairs aggregated from open-source datasets (TextCaps\footnote{\texttt{https://huggingface.co/datasets/HuggingFaceM4/the\_cauldron}, configuration \texttt{textcaps}.}, Localised Narratives\footnote{\texttt{https://huggingface.co/datasets/HuggingFaceM4/LocalizedNarratives}.}, COCO Captions\footnote{\texttt{https://huggingface.co/datasets/lmms-lab/COCO-Caption}.}, and subsets of Pexels/In\-ternVL captions\footnote{\texttt{https://huggingface.co/datasets/CaptionEmporium/pexels-568k-internvl2}.}, Conceptual Captions\footnote{\texttt{https://huggingface.co/datasets/google-research-datasets/conceptual\_captions}.}, and DataComp\footnote{\texttt{https://huggingface.co/datasets/mlfoundations/DataComp-12M}.}).
The training procedure samples from roughly one million image-caption pairs.

\begin{figure}[!h]
    \centering
    \includegraphics[width=0.99\linewidth]{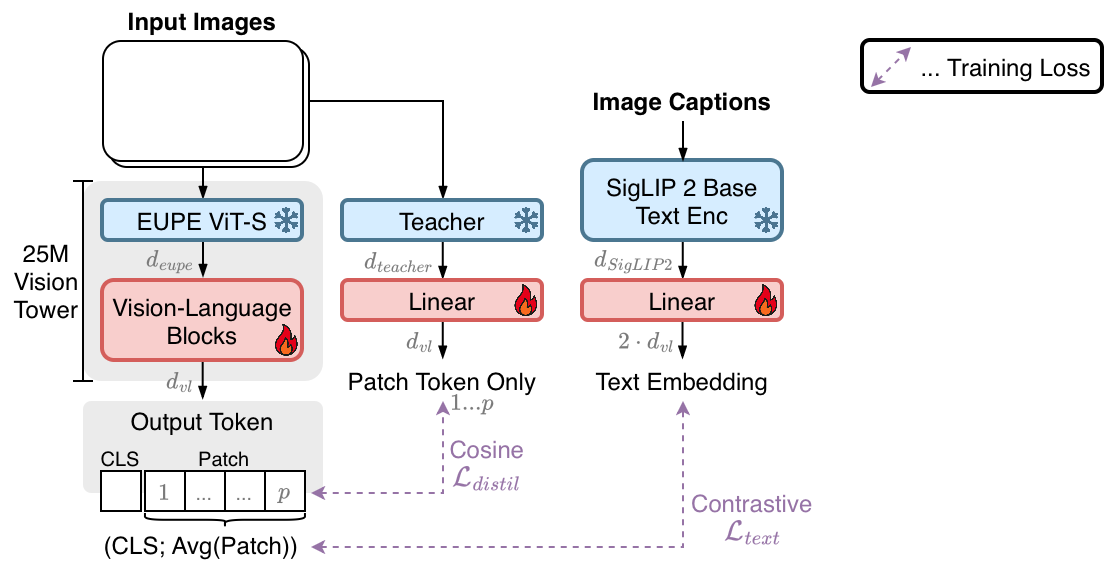}
    \caption{EUPE language-head training. A frozen EUPE ViT-S backbone provides CLS and patch tokens, which are mapped by a lightweight dino.txt-style head into a SigLIP~2-aligned language space. The head is trained with image-text contrastive supervision from a frozen SigLIP~2 text encoder and optional patch-level distillation from a frozen RadSeg teacher, yielding dense text-queryable patch embeddings from a compact, 25M parameter vision tower.}
    \label{fig:eupe_train}
\end{figure}

Fig.~\ref{fig:eupe_train} summarises the training protocol.
Captions are encoded with a frozen SigLIP~2 text encoder.
Captions and vision tower outputs are linearly projected to a shared feature dimension of $d_{vl}=512$.
The head is optimised with a bidirectional image-text contrastive loss $\mathcal{L}_{text}$ over in-batch negatives.
Following \cite{Jose2025DinoTXT}, the loss is applied over the CLS token and an average of the patch token.
Early training additionally applies a small patch-level distillation term from a frozen RadSeg \cite{alama2025radseg} teacher to encourage language-aligned patch features.
RadSeg builds on the RADIOv3 backbone using a modified SigLIP~2 head, providing patch features grounded in the SigLIP~2 vision-language feature space.  
This follows recent agglomerative distillation methods \cite{Ranzinger2024AMRADIO} that use separate heads per teacher.
We project dense teacher features to the vision tower's output resolution using a learned linear projection acting as a very small distillation head.
The distillation loss $\mathcal{L}_{distil}$ is calculated from cosine similarities between patch outputs and projected teacher outputs. 
Both contrastive and distillation losses are warmed up over the first epochs.
Training uses $512{\times}512$ crops, batch size $48$, AdamW with cosine learning-rate decay and mixed precision.

\textbf{Language Head Architecture.}
The language head follows the vision-language tower design in \cite{Jose2025DinoTXT}.
EUPE provides a normalised CLS token and a grid of normalised patch tokens.
These tokens are concatenated and passed through two transformer blocks with 8 attention heads, LayerNorm, residual self-attention, and an MLP with expansion factor $4$.
Stochastic depth with drop-path probability $0.3$ is used during training.
After the transformer blocks, a final LayerNorm and linear projection map all tokens to $d_{vl}=512$.

The transformed CLS token represents the image-level branch, while the transformed patch tokens form the dense language-grounded feature map.
For contrastive training, the global image representation is obtained by concatenating the transformed CLS token with the mean of the transformed patch tokens.
This yields a $2d_{vl}$-dimensional image representation that combines global image evidence and average patch-level evidence.
The frozen SigLIP~2 text encoder produces pooled text features, which are projected with a learned linear layer into the same $2d_{vl}$-dimensional space.
For dense retrieval at inference time, only the patch-aligned half of this text representation is used, matching the $d_{vl}$-dimensional patch tokens.
Thus, the same trained head provides both global image-text alignment for training and dense text-queryable patch embeddings for ReSiReg.

Training is implemented using PyTorch \cite{Ansel2024PyTorch} and run on two Quadro RTX 6000 GPUs.
Tab.~\ref{tab:eupe_training_hyperparameters} summarises training hyperparameters.

\begin{table}[!h]
    \centering
    \small
    \begin{tabular}{lcc}
        \toprule
        \textbf{Hyperparameter} & \textbf{Training Stage 1} & \textbf{Training Stage 2} \\
        \midrule
        Epochs & 100 & 350 \\
        Batch size & 48 & 48 \\
        Epoch length & 1000 & 1000 \\
        Optimiser & AdamW & AdamW \\
        Base learning rate & $6{\times}10^{-4}$ & $4{\times}10^{-4}$ \\
        Minimum learning rate & $1{\times}10^{-6}$ & $5{\times}10^{-6}$ \\
        Weight decay & 0.05 & 0.05 \\
        Learning-rate schedule & cosine & cosine \\
        Contrastive loss scale & $1.0$ & $1.0$ \\
        Contrastive warmup & 4 epochs & 4 epochs \\
        Patch distillation loss scale & $0.1$ & $0$ (disabled) \\
        Patch distillation warmup & 8 epochs & disabled \\
        Logit scale initialisation & $1/0.07$ & $1/0.07$ \\
        Maximum logit scale & 100 & 100 \\
        Image-caption training pairs & $\sim$600k & $\sim$1M \\
        \bottomrule
    \end{tabular}
    \vspace{2mm}
    \caption{Hyperparameters for the two-stage EUPE language-head training procedure. Stage~1 trains the head with bidirectional image-text contrastive learning and RadSeg patch distillation to bootstrap dense vision-language alignment. Stage~2 continues contrastive post-training on a larger dataset mixture without patch distillation to broaden language coverage.}
    \label{tab:eupe_training_hyperparameters}
\end{table}

\end{document}